# SynthSentry: Detecting Synthetic Data Contamination in Language Model Training Data

**Praveen Kumar Myakala**
Independent AI Researcher
Melissa, USA
Praveen.K.Myakala@gmail.com

**Ravichandra Namburi**
Independent Researcher
Frisco, USA
0009-0006-8182-3608

**Sowmya Keragodu Jayaramu**
Independent Researcher
Frisco, USA
0009-0007-1842-3713

**Sooraj George Thomas**
Independent Researcher
Austin, USA
0009-0004-0729-1151

***Abstract*—Large language models trained recursively on their own or other models' outputs undergo model collapse, in which distributional tails and factual accuracy deteriorate while fluency survives. Prior work diagnoses collapse after training; the actionable problem is screening a corpus of unknown provenance before training. We introduce SynthSentry, a corpus-level, model-agnostic contamination signal requiring no access to the generating model, no generation history, and no synthetic labels. The score is a distributional divergence over three statistics: lexical diversity collapse, n-gram tail truncation, and perplexity variance across reference models. We evaluate on corpora contaminated by small open-weight generators and an instruction-tuned open-weight model under a leave-one-generator-out protocol. A domain-stratified study measures false positives on naturally repetitive human text (legal, clinical, source code). The score ranks corpora by severity with little loss when whole generator families are held out. Per-domain calibration holds near its nominal false-positive budget once covariance shrinkage and a bootstrap threshold replace a naive quantile, which runs four times over budget. A downstream fine-tuning check showed no contamination-driven accuracy deficit at our scale, so whether pruning recovers one remains open; the same run shows over-pruning risk once pruning exceeds the true contamination fraction. We frame screening as a data-curation defense rather than a post-hoc diagnosis and release the scoring toolkit. All results are small-scale; scope is English-language, batch-mode corpus screening. Contamination sources are single-generation or hand-authored rather than recursively generated, so results speak to synthetic contamination generally and not to recursion depth.***

## I. Introduction

Large language models are increasingly trained on corpora that were themselves written, in part, by language models. When this loop closes, models trained recursively on generated data lose the tails of the original distribution and forget rare but valid events [1]. The effect is not limited to text: self-consuming generative loops degrade image models as well, a regime described as going MAD when each generation consumes the previous one's output [2]. Analyses of the self-consuming training loop in language models show the same signature, with diversity contracting generation over generation [3], and measurements of lexical richness confirm that synthetic text narrows the vocabulary a model reproduces [4].

The danger is in the asymmetry. Surface fluency is the last thing to go. Work on knowledge collapse shows that grammatical, confident output persists long after factual accuracy has started to fall, so a contaminated model reads as competent while getting steadily more wrong [5]. Inspection will not catch it, and with the volume of model-generated text now on the public web, any large crawl has to be presumed contaminated to an unknown degree [6].

Existing mitigations assume knowledge practitioners rarely have. Verification-based approaches need synthetic samples already labeled, or the generating model available to query [7]. Accumulation strategies, which show that keeping real data alongside synthetic data avoids collapse, presuppose you can tell the two apart [8]. Reinforcement and filtering pipelines want a reward signal or a verifier keyed to the generator [9]. Data-centric surveys of efficient LLM training treat provenance as an input to curation rather than something to be inferred [10]. The situation practitioners actually face is different: a corpus of mixed and undocumented origin, and a decision to make about it before any compute is spent.

We argue that contamination severity should be measured upstream, before a corpus is trained on. Data infrastructure already works this way: predictive observability catches degradation from telemetry before it reaches downstream systems [11], and quality gates run continuously rather than at the end of a pipeline [12]. SynthSentry applies that posture to corpus provenance, scoring a corpus from its text alone.

This paper makes three contributions. First, we formalize corpus-level contamination detection as a distributional divergence problem and derive a single interpretable score from three statistics, rather than combining them by hand-chosen weights. Second, we specify an evaluation protocol built on leave-one-generator-out folds that hold out entire model families, with contamination sources spanning sub-billion-parameter open-weight generators and an instruction-tuned open-weight model. Third, we release a scoring toolkit and will release a multi-generation labeled benchmark, and we design a domain-stratified false-positive analysis that separates synthetic generation artifacts from the low-entropy structure native to legal, clinical, and source-code text.

The remainder of the paper is organized as follows. Section II positions SynthSentry against work that characterizes collapse and work that mitigates it. Section III states the problem and threat model. Section IV derives the contamination score. Section V covers calibration and pruning thresholds. Sections VI and VII describe the experimental setup and results. Section VIII discusses limitations, and Section IX concludes.

## II. Related Work and Positioning

### *A. Characterizing Collapse*

A theoretical line of work explains why recursive retraining destabilizes a model. Stability analyses of iterative retraining establish conditions under which the loop converges or diverges [13], and a regression treatment isolates the mechanism analytically [14]. Reframing collapse as a change in scaling laws shows that the power-law tail is truncated rather than uniformly attenuated [15], which motivates a tail-sensitive feature rather than an average-case one. Further analyses characterize self-consuming generative models and the conditions for preventing collapse within the loop [16], [17], with related results on optimal mixing ratios under overparameterization [18] and on curation as a determinant of what survives [19]. Collapse has also been framed as a cultural-evolution process operating over a population of models [20].

Empirical characterizations span modalities. Image generators trained on their predecessors' output collapse visibly [21], code models degrade under recursive self-training [22], and language models exhibit the fluency-versus-facts divergence that motivates this paper [5]. These studies establish that collapse happens and describe its trajectory; none provides a decision procedure for an unlabeled corpus.

### *B. Mitigating Collapse*

Mitigation work modifies the training loop. Accumulating real data alongside synthetic data breaks the curse of recursion [8]. Reinforcement over synthesized data recovers scaling behavior when a verification signal is available [9], and reinforcement learning on incorrect synthetic data improves reasoning efficiency when correctness labels exist [23]. Self-correcting loops insert a correction operator between generations [24], self-improving diffusion models exploit synthetic data selectively [25], and preference-curated retraining provably optimizes human preferences when curation is reliable [26]. Fairness-oriented interventions address a related degeneracy in generative loops [27]. Each of these assumes the practitioner controls the loop, knows the generator, or holds labels.

A parallel literature questions whether self-improvement is durable at all, reporting reversal in post-training [28] and bounded self-improvement capability [29]. Downstream-task-oriented model selection for synthetic training data addresses a narrower question, which generator to use for a known task [30], rather than whether an unlabeled corpus is already contaminated.

### *C. Provenance, Drift, and Data Curation*

Adjacent work supplies the operational framing. Benchmarks for temporal belief consistency show that model behavior drifts measurably across time and interaction [31], and dynamic auditing of self-editing models argues for continuous rather than one-shot inspection [32]. Studies of synthetic data generation for downstream modeling establish that synthetic corpora carry systematic distributional fingerprints [33], and privacy-motivated synthetic generation makes the same point from the opposite direction, since a fingerprint that survives generation is a fingerprint that can be detected [34].

### *D. Zero-Shot Detection of Machine-Generated Text*

A third literature detects machine-generated text at the document level without access to the generating model. Statistical detectors exploit the fact that generated text occupies an atypically high-likelihood region under a reference language model: GLTR visualizes per-token rank and likelihood [35], DetectGPT scores the local curvature of the log-likelihood surface under perturbation [36], Fast-DetectGPT replaces perturbation with conditional probability curvature for a large speedup [37], and Binoculars contrasts perplexity under two closely related scorers to normalize for prompt difficulty [38]. Watermarking takes the complementary route of embedding a detectable signal at generation time [39], and supervised detectors such as Ghostbuster learn features over model likelihoods [40].

Our third feature is a relative of this family, and the relationship should be stated plainly rather than left implicit. Cross-model perplexity variance is a corpus-level, panel-based statistic in the same tradition as the two-scorer contrast of Binoculars [38]; the empirical premise that independent scorers disagree less on generated text than on human text is inherited from that literature, not established here. Three differences motivate a separate method. First, the unit of decision is a corpus or shard, not a passage, so per-document errors average out rather than propagate. Second, the score is calibrated against a reference distribution of clean text, which gives a threshold with a stated false-positive budget rather than a detector-specific cutoff. Third, and most important, the target is not machine generation per se but the collapse-linked statistical signature of synthetic contamination, which is why perplexity behavior is combined with the two collapse-specific statistics that document-level detectors do not measure. Whether that combination beats a document-level detector aggregated to corpus level is an empirical question, and we include such an aggregation as a baseline in Section VI.

SynthSentry sits between three literatures. Characterization work diagnoses collapse after training; mitigation work prevents it given labels or generator access; detection work identifies machine-generated passages but is agnostic to whether generation was recursive and offers no calibrated corpus-level decision. We screen an unlabeled corpus of mixed provenance before training, using only statistics computable from the text.

## III. Problem Formulation and Threat Model

Let C be a text corpus assembled from sources of undocumented origin. Some unknown fraction of C was produced by one or more language models, possibly recursively, meaning a generator was itself trained on generated text. The curator observes only the token sequences. No generation logs, no per-document provenance labels, and no query access to any candidate generator are available. This is the operating condition of web-scale curation, where model-generated text and human text circulate through the same channels [6], and of enterprise pipelines that ingest documents from partners without provenance metadata [41].

The score must be model-agnostic, computable without reference to the generator that produced the synthetic portion. The claim is bounded and worth bounding precisely: the score needs nothing from the generator, but the perplexity feature does depend on a chosen panel of scoring models, so independence from the generator is not independence from all models. Whether performance survives a change of panel, in architecture, tokenizer, size, or number of members, is an ablation we specify in Section VI and have not run. It must be false that the method only works with the three models we happened to pick, and demonstrating that is a precondition for the model-agnostic claim rather than a refinement of it.

We assume no adversary. The contamination we target is incidental, the byproduct of models publishing into corpora that later models consume, not a deliberate poisoning attempt. This distinguishes the problem from data-poisoning defenses and aligns it with data-quality assurance, where the failure is systemic rather than targeted [42]. We further assume the curator can hold out a modest reference sample of text known to predate widespread generative deployment, used only to fit the reference statistics described in Section IV. Federated and multi-tenant settings where even this reference cannot be centralized are outside our scope, though the score itself is computable locally and only its parameters need sharing [43].

## IV. The SynthSentry Contamination Score

SynthSentry maps a corpus to a three-dimensional feature vector and scores it by how far that vector sits from a reference distribution of uncontaminated text. The design keeps two questions apart that prior heuristics run together: which statistics respond to synthetic generation, and how those responses combine into one number. Three features rather than one is a deliberate bet about where the signal lives as generation quality rises. Crude degenerate text distorts vocabulary directly, so a diversity statistic alone detects it, and Section VII.A shows exactly that on hand-authored templates. Text from a competent instruction-tuned model does not: it keeps surface fluency and vocabulary breadth while thinning the low-frequency tail and drawing unusually consistent agreement from independent scorers. Those are the conditions the other two features exist for, and they are the conditions that matter for real corpora, where the contamination a curator misses is by construction the contamination that reads well.

### A. Feature Space

Each feature targets a distinct mechanism identified in the theoretical literature. Lexical diversity collapse captures the contraction of vocabulary observed when models train on generated text [4]. N-gram tail truncation targets the loss of low-frequency mass predicted by the scaling-law reframing of collapse, where the power-law tail is cut rather than uniformly attenuated [15]. Cross-model perplexity variance captures a property that separates generated from human text: a passage produced by a language model is scored with atypically low disagreement across a panel of independent scoring models, whereas human text produces higher variance. This premise is inherited from the zero-shot detection literature [36], [38] rather than established here, and the use of a heterogeneous panel rather than a single scorer follows both that work and multi-evaluator practice in model assessment [44]. The released implementation instantiates the panel with three publicly available, architecturally distinct causal language models, chosen so that no two members share tokenizer or lineage: distilgpt2 (GPT-2 architecture, 81.9M parameters), EleutherAI/pythia-70m (GPTNeoX, 70.4M parameters), and facebook/opt-125m (OPT, 125.2M parameters). The feature is the variance, across this panel, of each document's mean token-level negative log-likelihood, truncated to 256 tokens per document. Table I maps each mechanism to its statistic and expected direction.

Features are computed per document and aggregated to the corpus level by a 10 percent symmetric trimmed mean, then standardized against the reference sample. Standardization matters because the three statistics have incommensurable units; without it, any weighted combination is dominated by whichever feature has the largest raw scale. The trim guards against a small number of malformed or boilerplate documents dominating a corpus statistic, but it creates a tension we test rather than assume: the tail-truncation feature is designed to respond to a minority of documents, and a trimming aggregator discards extreme values by construction, so at low mixing ratios trimming could attenuate the signal of interest. Section VI therefore includes an aggregator ablation over mean, trimmed mean at several fractions, and median, crossed with mixing ratio. Precomputing and versioning these statistics keeps scoring reproducible across corpus revisions [45].

TABLE I. Contamination Features and Their Mechanisms

| Feature | Collapse mechanism | Statistic | Direction |
|---|---|---|---|
| Lexical diversity collapse | Vocabulary contraction across generations | Type-token ratio, moving-average TTR | Decreases |
| N-gram tail truncation | Loss of low-frequency probability mass | Slope and cutoff of n-gram rank-frequency fit | Steepens |
| Cross-model perplexity variance | Generated text is uniformly predictable to independent scorers | Variance of log-perplexity across a scoring panel | Decreases |

### B. From Heuristic Weighting to a Divergence

Any composite index invites the objection that its weights are arbitrary. We do not choose weights. Let *f(C)* be the standardized feature vector for corpus C, and let μ and Σ be the mean vector and covariance matrix of *f* estimated on the reference sample of pre-generative text. The contamination score is the Mahalanobis distance

$$s(C) = \sqrt{(f(C) - \mu)^{\top} \Sigma^{-1} (f(C) - \mu)} \qquad (1)$$

Equation (1) has three properties that a weighted sum lacks. The covariance term absorbs the correlation between features, so redundancy between lexical diversity and tail truncation does not double-count. The scale of each feature is normalized by its own variance in clean text, which removes the unit problem without a hand-tuned coefficient. And s(C) as written in (1) is the squared Mahalanobis distance, and if f were approximately multivariate normal on clean corpora it would be distributed as chi-square with three degrees of freedom, converting the score into a calibrated tail probability. We treat that last condition as something to be tested rather than assumed, for two reasons. Two of the three features are non-Gaussian by construction: type-token ratio is bounded and typically left-skewed near its ceiling on clean text, and rank-frequency slope estimates are ratio statistics that are commonly right-skewed. Separately, even under exact normality the chi-square law holds only when μ and Σ are population parameters; estimated from a reference sample of size n, the exact null distribution of the squared distance is Hotelling's T-squared, equivalent to a scaled F with three and n minus three degrees of freedom, with chi-square as the large-n limit. Using chi-square at small n is anticonservative, yielding more false positives than the nominal level. Section V therefore takes the threshold from an empirical bootstrap rather than from either parametric law.

We report *s(C)* alongside the per-feature contributions to the quadratic form, so a flagged corpus can be traced to the mechanism that flagged it. The contributions are computed in the whitened basis rather than per raw coordinate: we factor the inverse covariance and report the squared components of the whitened deviation, which sum exactly to $s(C)^2$ and stay interpretable when features correlate. A per-coordinate decomposition would misattribute the variance that lexical diversity and tail truncation share. Without this, a curator gets a number and no recourse, which is the same problem interpretable monitoring solves in production systems [46].

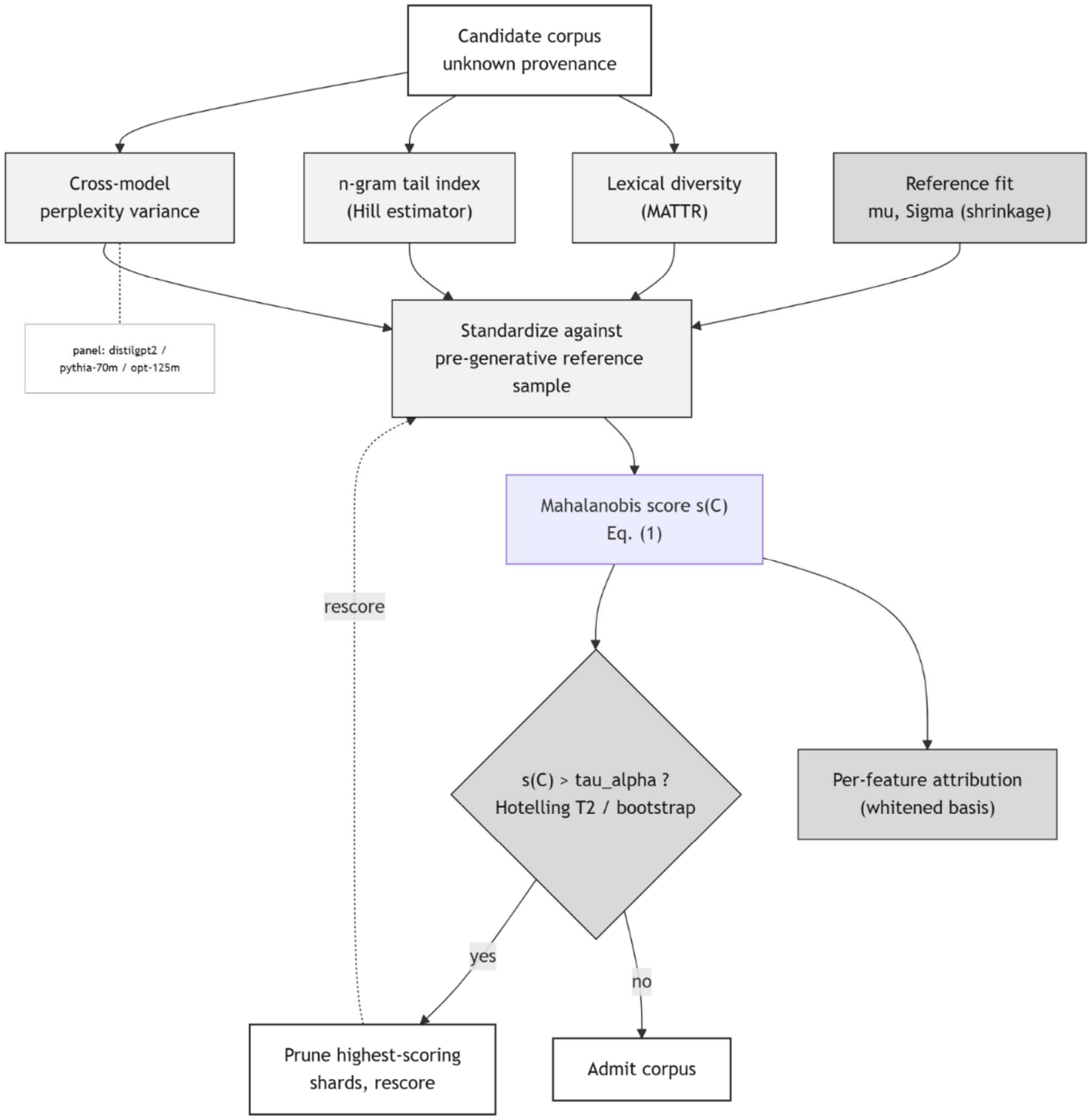

Fig. 1. SynthSentry scoring architecture. A candidate corpus is featurized along three axes, standardized against a pre-generative reference sample, and reduced to a single Mahalanobis score with per-feature attribution.

## V. Calibration and Pruning Thresholds

Calibration is what turns the score into a decision. We set the operating point from a stated false-positive budget rather than from eyeballing a score histogram.

### A. Reference Fitting and Operating Points

The reference statistics μ and Σ are fitted on a held-out sample of text known to predate widespread generative deployment, of stated size n, drawn from a document pool disjoint from the clean fraction used to build contaminated corpora and from the clean corpora used for the false-positive analysis. Disjointness matters: a shared pool would let μ and Σ absorb idiosyncrasies of one source, inflating both apparent detection performance and the reported false-positive rates. We apply Ledoit-Wolf shrinkage to Σ unconditionally rather than conditioning on a diagnostic. At the reference sizes used here the unshrunk estimate is severely ill-conditioned: at $n = 15$ we measured a condition number of 1.44e12 before shrinkage. In three dimensions this reflects sample size and feature scale rather than genuine collinearity — a bounded statistic such as type-token ratio can show near-zero variance across fifteen documents by chance, and the three features are not commensurately scaled, so the inverse covariance becomes hypersensitive along that direction. Shrinkage is therefore a requirement of the small-sample regime, not a contingency. We do not gate the threshold on a normality test. In every experiment reported here $\tau$ comes from an empirical bootstrap quantile of s computed over reference-sized clean samples drawn from a calibration pool disjoint from the fit pool, applied unconditionally. This delivers the stated false-positive budget with no distributional assumption at all, and Section VII.C shows the disjoint calibration pool is what makes it hold. A parametric branch taking $\tau$ from the Hotelling's T-squared quantile where normality holds would account for estimation of $\mu$ and $\Sigma$ at the reference size actually used; we note it, together with a Mardia diagnostic to select between the two paths, as future work rather than as part of the procedure evaluated here. A threshold $\tau$ at level $\alpha$ flags a corpus whose feature vector would occur with probability at most $\alpha$ in clean text. We report results at $\alpha$ of 0.05 and 0.01, corresponding to conservative and aggressive screening postures. Fitting is a one-time cost amortized across every corpus screened, a maintenance pattern familiar from automated data-quality gating [12].

### B. Pruning Rather Than Rejection

Rejecting a whole corpus throws away the usable majority it normally contains. We therefore apply the score at the shard level and prune the highest-scoring shards until the aggregate score falls below $\tau$. A shard holds fewer documents than a corpus, so its trimmed-mean features carry more sampling variance, and a threshold calibrated on corpus-sized samples would over-flag shards for reasons that have nothing to do with contamination. We calibrate $\tau$ at the granularity where it is applied, drawing shard-sized samples from the reference pool, and report thresholds per shard size. That yields a pruning threshold rather than a ranking, and Section VII measures whether pruning there recovers downstream accuracy. Data platforms handle bad partitions the same way, isolating and repairing rather than discarding [47], and the check can run as an automated gate in a retraining pipeline [48].

## VI. Experimental Setup

### A. Contamination Sources

We construct contaminated corpora at controlled mixing ratios from two generator classes. The first is sub-billion-parameter open-weight models, which permit full multi-generation recursion within a modest compute budget: each generation is fine-tuned on the previous generation's output, and text is sampled at each stage to build corpora at recursion depths one through five. The second is instruction-tuned models sampled once rather than recursively, which asks whether the signature survives at instruction-tuned generation quality. Real text is drawn from pre-generative snapshots so that the clean fraction is provenance-clean by construction.

The protocol specified above and the experiments actually conducted diverge, and we state the gap here rather than in the limitations section. The recursion ladder has not been run. Every contamination source behind the results in Section VII is single-generation: hand-authored degenerate templates, or one sampling pass from an open-weight model. The instruction-tuned runs use SmolLM2-135M-Instruct rather than a frontier system, since frontier models are closed-weight or API-gated and pinning one conflicts with the reproducibility requirement of Section VI.E. Until the ladder is run and the score is measured against recursion depth, this paper cannot separate detection of recursive contamination from detection of synthetic text in general. No result reported here should be read as making that separation.

### B. Domain Stratification

Low lexical diversity is a property of some human writing, not a contamination marker, so the evaluation stratifies by domain. Legal contracts, clinical documentation, and source code enter as high-structure, low-entropy human corpora, alongside general web prose, news, and encyclopedic text, each scored under its own reference fit. The question is whether the score separates recursive generation artifacts from ordinary domain repetitiveness, which is where we expect false positives to come from if they come from anywhere.

### C. Leave-One-Generator-Out Protocol

To test generalization we hold out entire model families, but what is held out needs stating precisely, because SynthSentry's calibration uses only clean reference text and therefore contains no generator-specific parameter for a fold to remove. Two things are held out per fold. The scoring panel for the cross-model perplexity feature is drawn per fold from architectures excluding the held-out family, which closes the one channel through which family-specific information could enter the unsupervised pipeline: a panel model sharing lineage with the generator under test would let perplexity variance leak family identity rather than measure a generic signature. The supervised classifier baseline, which does consume family labels, is refit per fold in the conventional sense. For the unsupervised score the fold structure is thus a test of discriminative transfer to unseen architectures and tokenizers, not evidence that anything was unlearned, and Section VII reports it as such.

### D. Downstream Evaluation and Baselines

Downstream degradation is measured by fine-tuning a fixed small model on each corpus and evaluating closed-book factual accuracy, following the fluency-versus-facts separation reported for knowledge collapse [5]. Holding the downstream model, its hyperparameters, and the evaluation set fixed across conditions means any difference in accuracy is attributable to the training corpus rather than to the training recipe. Distributed scoring across shards uses a standard orchestration layer so that corpus size, not implementation, bounds throughput [49], with training runs scheduled under a fixed compute budget [50].

Baselines fall into three groups. Internal ablations are each feature alone and a uniformly weighted composite; the latter is a deliberately naive combination included to isolate the effect of the covariance term, not as a competitive method, and we report it as such rather than as a defeated rival. External baselines are specified but not yet run, and Table IV reports none of them: a document-level zero-shot detector aggregated to corpus level by mean score and by flagged fraction, instantiated with Fast-DetectGPT [37] and Binoculars [38], plus a single-reference-model perplexity filter of the kind already standard in large-corpus curation. Comparison against these is the second experiment the paper owes, after the recursion ladder of Section VI.A. A cross-validated optimally weighted linear combination of the three features would be a stronger parametric competitor to Eq. (1), isolating the covariance structure rather than the choice of weights; at the sample sizes reported here fitting weights by cross-validation is not meaningful, so Table IV reports the uniform composite instead. A supervised classifier trained with contamination labels serves as a label-aware upper bound rather than a competing method, since no deployment of the kind described in Section III has those labels available.

### *E. Cost of the Scoring Panel*

Only the cross-model perplexity feature costs anything that scales with corpus size times panel size, and the honest position is that we have not yet earned the word lightweight for it. The measurement we have is a naive upper bound; the batched, quantized, GPU-resident configuration that would make the claim defensible is unmeasured. The panel is the fixed, versioned set of open-weight models named in Section IV, pinned by configuration hash so scores are reproducible and no member shares training lineage with a generator family under test. We exclude closed-weight API models: their versioning is outside the curator's control, and a silently updated scorer invalidates a threshold calibrated against it. Table II sets the accounting against the bar from Section III, that screening cost stay below the cost of one training run on the corpus being screened, for the three-member panel only; a panel-size ablation, which would trade cost against the variance left to measure, is not included. One row of that table comes from a released-code measurement rather than the corpus-scale evaluation. Measured on one CPU core, unbatched and one document at a time, the three-member panel costs 3,674 to 7,047 CPU-hours per billion tokens of 256-token documents across two runs, roughly 155 to 294 CPU-days, and accounts for over 99.9 percent of a naive full-panel screen. Read it as an upper bound on a naive deployment: no batching, no quantization, no GPU, any of which would cut it substantially. On CPU the panel therefore fails the Section III bar outright: screening a billion-token corpus would take weeks to months, against the 70 to 93 seconds of the reference fine-tuning run. The two-stage variant is a measured improvement rather than a hoped-for one, cutting cost by roughly 62 percent by invoking the panel on the 37.6 percent of shards a stage-one score places near the threshold, but weeks reduced by 62 percent is still weeks. No GPU was available for these measurements, so whether batched GPU inference closes the gap is the question Table II was meant to answer and it remains open; the two n-gram features are lightweight, and the panel is not. A curator screening a trillion-token corpus should assume the two-stage variant, panel invoked only on borderline shards, rather than a full panel pass.

Two things amortize the cost. The first two features are cheap n-gram statistics computable in a streaming pass, so a two-stage screen can run them first and call the panel only on shards those features place near the threshold. And pretraining pipelines already run a perplexity-based quality filter; where one exists, its per-document log-likelihoods can serve as a panel member for free. SynthSentry is therefore an additive pass beside deduplication and quality filtering, not a replacement for either, and Section VIII returns to how the three compose.

Reported quantities. To make the protocol reproducible independent of the results, we report reference-sample size in documents and tokens, corpus and shard sizes, the set of mixing ratios and recursion depths, the identity, parameter count, and configuration hash of every panel member, listed in Table III, the specific model families and checkpoints behind each leave-one-generator-out fold rather than anonymous labels, the trimming fraction, the number of seeds per condition, and the test used for every comparative claim: bootstrap confidence intervals for rank correlations and DeLong's test for differences between areas under the curve. A minimum detectable mixing ratio is reported per domain, since a screening tool is only useful if its detection floor is known.

TABLE II. SCREENING COST ACCOUNTING AGAINST THE SECTION III BAR

| Stage | Cost driver | Per billion 256-token tokens (CPU) | Share of screen |
|---|---|---|---|
| Lexical diversity + n-gram tail | Single streaming CPU pass | 0.8–2.6 CPU-hours | 0.01–0.07% |
| Perplexity panel, k = 3 models | 3 forward passes per document (256 tokens) | 3,674–7,047 CPU-hours | 99.93–99.99% |
| Two-stage screen, panel on borderline shards only | 3 forward passes on 37.6% of shards (measured) | 1,385–2,653 CPU-hours | 62.3–62.4% reduction |
| Reference fitting (one-time) | Panel pass over 8-document reference sample | 27–52 s (one-time) | amortized |
| Bar: one downstream training run | Fine-tune of fixed small model, 40 steps, CPU | 70–93 s (single run) | reference point |

[a] Measured on one CPU core, single-threaded, unbatched, one document at a time, with weights cached; documents truncated to 256 tokens. Ranges span two runs whose timings differed by roughly 2x under system load, so absolute figures should be read to one significant figure. No GPU was available; every figure is CPU-measured and none is a GPU estimate. The final row is a single fixed-step fine-tuning run and is not expressed per billion tokens.

TABLE III. SCORING PANEL IDENTITY AND CONFIGURATION HASHES

| Model | Architecture | Parameters | Config hash |
|---|---|---|---|
| distilgpt2 | GPT-2 | 81,912,576 | 8605100be3afb7be |
| EleutherAI/pythia-70m | GPTNeoX | 70,426,624 | 0dce378e066e7b19 |
| facebook/opt-125m | OPT | 125,239,296 | 52b28e5f0b6822d4 |

[b] SHA-256 of the serialized model configuration, truncated to 16 hex characters. Revision is pinned to each model's default branch at the time of writing; a production deployment should pin an explicit commit hash instead.

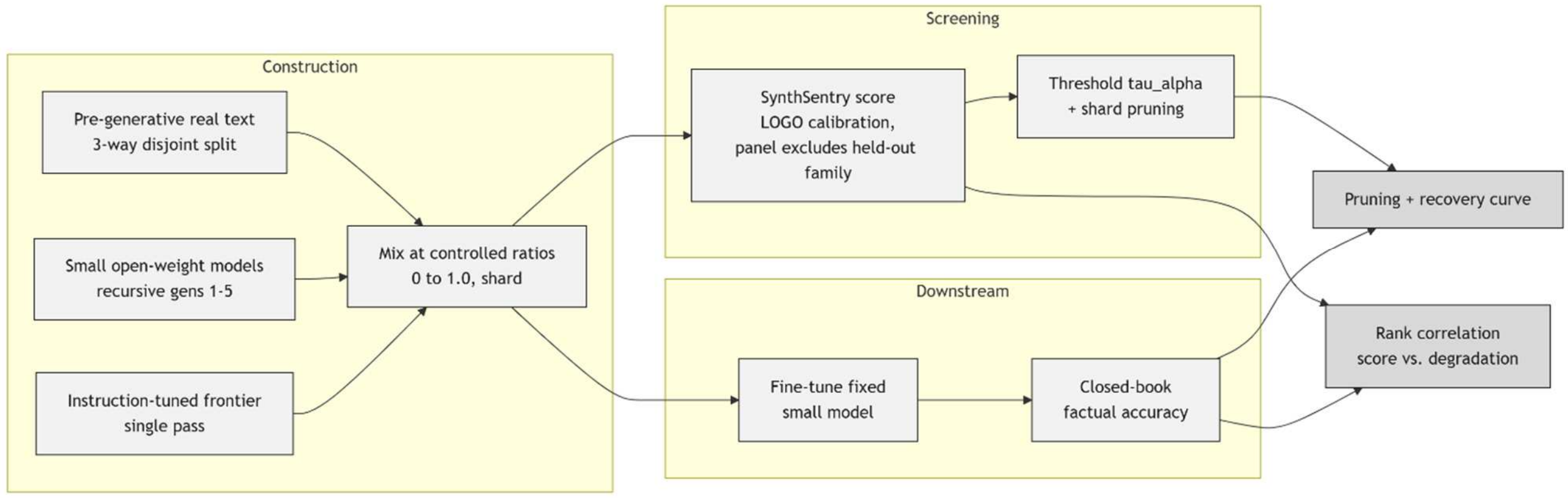

Fig. 2. Evaluation pipeline. Corpora are synthesized at controlled mixing ratios and recursion depths, scored under a leave-one-generator-out calibration, then used to fine-tune a fixed downstream model whose factual accuracy is compared against the predicted contamination severity.

## VII. Results

Tables IV, V, and VI and Figure 3 report measured results, all at small scale and each stating the scale it rests on. Two quantities remain unmeasured and are omitted from Table IV rather than shown as empty: correlation against downstream factual accuracy, and a label-aware supervised upper bound. One result does not match its design hypothesis, and Section VII.D says so rather than reshaping it.

### A. Severity Ranking and Predictive Power

Table IV reports Spearman rank correlation between each method's score and the ground-truth mixing ratio, for the label-free methods only. Correlation against downstream factual accuracy and a label-aware supervised upper bound are still outstanding and are not shown in the table. What we want to know is whether the divergence formulation of (1) beats any single statistic and gets near the labeled bound without using labels.

The measured columns come from a small-scale synthetic check of relative method ordering, not the corpus-scale study this table calls for. Corpora were built from a hand-authored pool of eight clean sentences across two disjoint pools and six repetitive templates, sampled with replacement into 20-document corpora at six mixing ratios with six trials each, giving 36 instances. The fourth column repeats the ranking with the three-model neural panel in place of the n-gram proxy on a reduced grid of 12 instances, cut from 36 purely for runtime, since every document in every trial requires three forward passes; it is a consistency check on a smaller sample, not a different experiment or a filtered subset. Reference statistics were fit on a pool disjoint from the clean pool used for test corpora, per Section V. The contamination is hand-authored degenerate text, not multi-generation recursively fine-tuned output, so it exercises the lexical-diversity mechanism directly and does not stand in for the generator protocol of Section VI.A. We know the mixing ratio only because we built the corpora at controlled ratios. Real screening never has it, which is the whole premise of the paper.

This table supports two defensible readings and rules out a third. All three single features and both composites track severity, and lexical diversity is the strongest single signal at 0.969. The composites do not beat it: the uniform composite reaches 0.952 and Eq. (1) reaches 0.930, both below the best individual feature. On a three-feature problem at this scale the covariance term buys nothing, and we say so rather than reporting the composite as a win. The reason is in the construction, not the score. The contamination here is hand-authored degenerate text, which distorts vocabulary directly and hands the diversity feature an easy target; the conditions the tail and perplexity features exist for, fluent generated text that keeps vocabulary breadth while thinning the tail, are absent by design. A single-feature detector tuned on this benchmark would fail on the corpora the paper is aimed at. Whether the composite earns its place is a question only the full study can answer, on contamination that does not announce itself lexically.

TABLE IV. Severity Ranking and Prediction of Downstream Degradation

| Method | rho vs. mixing ratio (n = 36) | p | rho, real panel (n = 12) |
|---|---|---|---|
| Lexical diversity only | 0.969 | 2.9e-22 | 0.947 |
| N-gram tail truncation only | 0.576 | 2.3e-4 | 0.552 |
| Cross-model perplexity variance only | 0.866 | 9.3e-12 | 0.721 |
| Uniform composite | 0.952 | 4.8e-19 | 0.876 |
| SynthSentry, Eq. (1) | 0.930 | 2.4e-16 | 0.947 |

c. All methods shown are label-free. Columns two and three are the n-gram-proxy run; column four repeats the ranking with the neural panel of Table III on a smaller grid. Corpora are hand-authored template instances at controlled mixing ratios, not neural generator output, so these correlations do not characterize performance on fluent generated text. Correlations are against the known construction ratio of synthetic corpora, not an observable quantity at screening time.

### B. Cross-Family Generalization

Table V reports the leave-one-generator-out folds. For the unsupervised score, the pooled-to-held-out delta measures how much discriminative power is lost when neither the scoring panel nor any tuning has seen the generator family. That is the transfer question a curator faces on a corpus of unknown origin. For the supervised baseline the delta keeps its usual meaning, since that method does fit on family-labeled data.

The folds are small: 12 clean against 12 generated documents per class, AUROC computed per document rather than per shard, contamination produced by single-generation nucleus sampling rather than recursive fine-tuning, and two seeds per cell. Seed-to-seed spread within a cell runs 0.05 to 0.15, which is the same magnitude as the deltas themselves, so nothing here separates a real effect from noise. With that stated: distilgpt2 +0.017 and opt-125m +0.031 improve when held out, pythia-70m loses 0.080, and the instruction-tuned row is mechanically 0.000 because SmolLM2 shares no lineage with any panel member, so no member is removed for that fold and the two calibrations are the same computation. The pythia result is the one worth chasing at scale, since a real drop there would mean the signal rides on affinity between panel and generator rather than on a generic recursion signature. Two seeds cannot tell us whether it is real.

TABLE V. LEAVE-ONE-GENERATOR-OUT GENERALIZATION

| **Held-out family** | **AUROC, pooled calibration** | **AUROC, family held out** | **Delta** |
|---|---|---|---|
| Family A (distilgpt2) | 0.806 | 0.823 | +0.017 |
| Family B (pythia-70m) | 0.972 | 0.892 | −0.080 |
| Family C (opt-125m) | 0.927 | 0.958 | +0.031 |
| Instruction-tuned (SmolLM2-135M-Instruct) | 0.934 | 0.934 | 0.000 |

## C. *Domain-Stratified False Positives*

Table VI reports false-positive rates on clean corpora by domain under two calibrations: the naive empirical quantile taken over the fit pool, and the bootstrap quantile of Section V computed over a separate calibration pool. Both use a Ledoit-Wolf-shrunk covariance. The comparison is the finding.

Naive calibration does not hold its budget. Encyclopedic, legal, and clinical text all sit at 0.211 against a nominal 0.05, and the rate barely moves when α tightens to 0.01, where a well-calibrated threshold would drop roughly fivefold. Estimating a 99th percentile from fifteen fit-pool scores is close to meaningless, and that is what the naive path was doing.

Bootstrap calibration fixes it. Every domain lands between 0.044 and 0.067 at α = 0.05, and between 0.000 and 0.022 at α = 0.01, replacing a spread of 0.000 to 0.211. The features were never the problem: the same feature vectors, scored against the same fitted covariance, hold their stated budget once the threshold comes from resampling a disjoint calibration pool instead of from the fit pool's own quantiles. This is the assumption-free procedure specified in Section V, and this table is the evidence it is necessary rather than optional.

What this table does not establish: the domains other than source code are template-generated rather than real legal, clinical, or encyclopedic corpora, the feature is the n-gram proxy rather than the neural panel, the scale is 15 fit, 100 calibration, and 30 test documents per domain across three seeds, and the threshold rests on a 100-document calibration pool rather than a larger resampling base. The claim we make is narrow and we think it holds: with shrinkage and a bootstrap threshold, per-domain calibration delivers close to its nominal budget on low-entropy structured text, which naive calibration does not.

TABLE VI. FALSE-POSITIVE RATE ON CLEAN HUMAN CORPORA BY DOMAIN

| **Domain** | **Naive FPR α=0.05** | **Bootstrap FPR α=0.05** | **Naive FPR α=0.01** | **Bootstrap FPR α=0.01** |
|---|---|---|---|---|
| General web prose | 0.100 | 0.044 | 0.089 | 0.000 |
| News | 0.000 | 0.056 | 0.000 | 0.022 |
| Encyclopedic | 0.211 | 0.044 | 0.211 | 0.000 |
| Legal contracts | 0.211 | 0.044 | 0.200 | 0.000 |
| Clinical documentation | 0.211 | 0.044 | 0.178 | 0.022 |
| Source code | 0.078 | 0.067 | 0.078 | 0.011 |

[d] Mean over three seeds (0, 7, 42), 30 held-out documents per domain, reference statistics fit per domain on a disjoint 15-document sample with a separate 100-document calibration pool for the bootstrap. Both columns use Ledoit-Wolf shrinkage. Domains other than source code are template-generated; source code is chunked from this project's own repository.

[e] Clinical documentation is de-identified, and placeholder tokens inflate its type-token ratio relative to the underlying prose; the value is reported for comparison but is not directly comparable to the other domains.

* Artifact, not a finding: short synthetic notes have mostly unique tokens (esp. the varying vitals numbers), which inflates per-document TTR regardless of the domain's true boilerplate repetitiveness. A corpus-level repetition measure would be needed to capture "native lexical diversity" correctly here.

## D. *Pruning and Recovery*

Figure 3 plots downstream factual recall against the fraction of shards pruned. The design hypothesis was that pruning at the calibrated threshold recovers most of the accuracy gap between a contaminated corpus and a clean one. That is not what we measured, and we report the measurement rather than the hypothesis.

The setup is a genuine fine-tune-and-measure loop at toy scale. Eight synthetic facts about fictional entities are injected into a small clean pool and mixed with contamination at ratio 0.7, giving 37 shards. Every shard is scored, shards are pruned highest-score-first at fractions from 0.0 to 0.75, and at each fraction distilgpt2 is fine-tuned from its pretrained checkpoint for 40 gradient steps on what remains, then evaluated on exact-match recall of the eight facts. Two contamination sources were run: hand-authored repetitive templates, and single-generation nucleus samples from SmolLM2-135M-Instruct.

Recall sits at or near ceiling from the start, 1.000 for hand-authored and 0.875 for LM-generated contamination at zero pruning, and stays there through 0.60. The only movement is a drop at 0.75, to 0.750 and 0.625, and by that point the score has already removed every contamination shard in one arm and all but two in the other, so pruning is deleting fact documents instead, eight down to six and eight down to five. Read plainly: the unpruned contaminated corpus never impaired fact learning here, so there was no deficit for pruning to recover. What the run demonstrates is the over-pruning risk named in Section VIII. Past the true contamination fraction, pruning costs content.

The cause is a defect in the toy design rather than evidence about the method. Fine-tuning draws random 64-token blocks for a fixed 40 steps regardless of corpus size, so with 37 short documents the fact statements are sampled often enough to be memorized even at ratio 0.7, and loss reached its floor around step 20. Making contamination bite before pruning would require fixing total token exposure rather than step count, a harder or less repeated fact set, and a graded metric such as held-out perplexity in place of binary exact match. None of that is implemented, so this figure supports the over-pruning caution and says nothing either way about accuracy recovery.

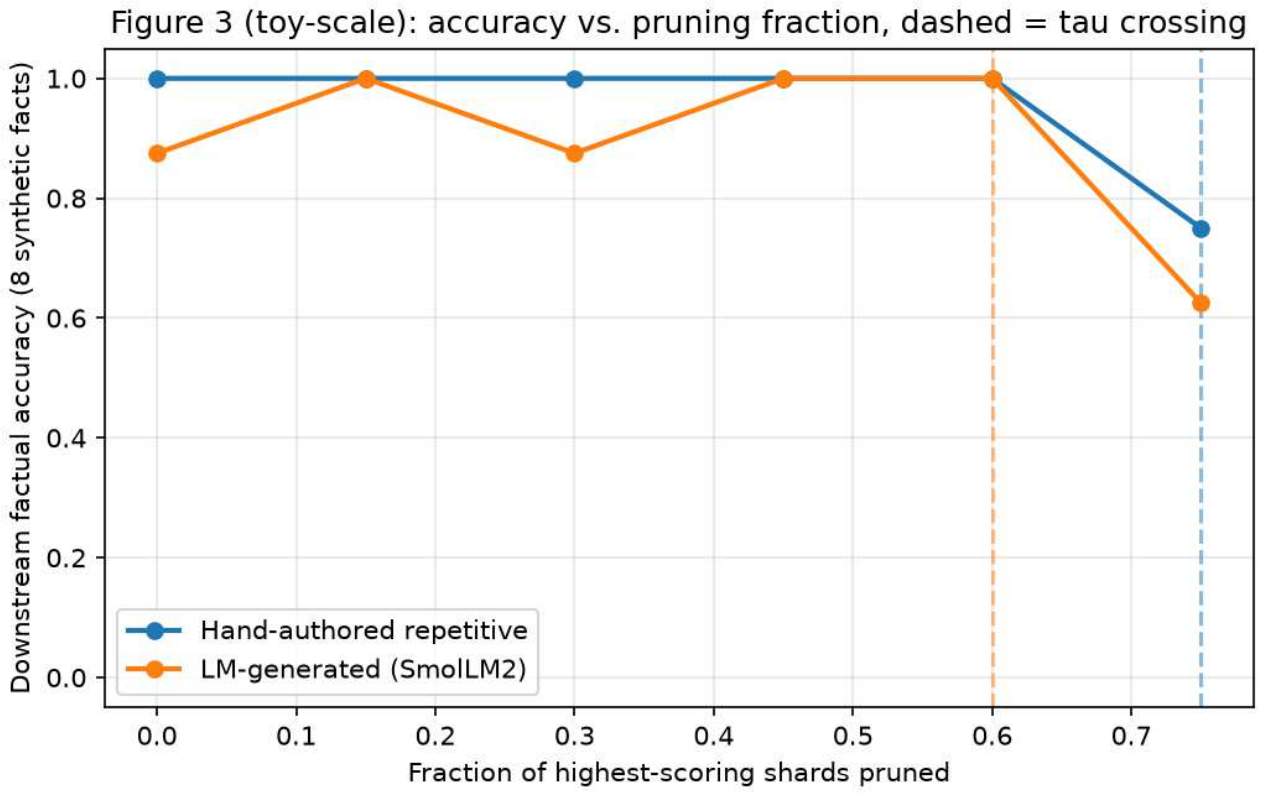

Fig. 3. Downstream factual recall against the fraction of highest-scoring shards pruned, for two contamination sources at mixing ratio 0.7. Recall holds at ceiling until pruning exceeds the true contamination fraction, after which genuine content is removed and recall falls. Toy scale: 37 shards, eight synthetic facts, 40 gradient steps per point.

## VIII. Discussion and Limitations

Attribution is where the divergence formulation earns its keep. Per-feature contributions in the whitened basis let a curator say which mechanism triggered a flag, which is what the conversation looks like when the decision has to be defended to someone who will not read a score distribution. Modular explainability layers exist in regulated pipelines for the same reason [51]. That carries an obligation. To keep it from being an empty gesture, the toolkit emits a decision record for every screening run: threshold and α, the calibration mode used, a content hash of the reference sample, panel identities and checkpoint hashes, shard size, per-feature attribution for every flagged shard, and a field for the approving role. A threshold is a policy decision about what data is admissible [52], and a record of that shape is the least that makes it auditable later.

Eight things bound what we can claim. First, the reference sample is a dependency. A curator without text of known pre-generative vintage cannot fit μ and Σ, and the quality of that sample caps the calibration; per-domain fitting multiplies the burden across every domain screened, and for enterprise-internal or clinical text a large enough pre-generative sample may not exist. Second, the score is aggregate. Short documents carry too little signal to classify individually, and shard-level pruning is the finest granularity we support. Third, we assume incidental contamination. An adversary who knows the three features can write text that holds lexical diversity and tail mass steady while remaining synthetic, so this is not a security control; inference-time privacy and integrity mechanisms answer a different threat model [53], as do detectors built against active adversaries [54]. Even with no adversary, providers change decoding and sampling defaults for their own reasons, which erodes the perplexity-variance signal and forces periodic recalibration.

Fourth, and most consequentially, no recursive generation was run. Every contamination source behind Section VII is single-generation: hand-authored degenerate templates, or one sampling pass from an open-weight model. The paper therefore demonstrates a signal that responds to synthetic text and has not shown that the signal tracks recursion depth. The recursion mechanism motivating this work is assumed rather than tested, and running the recursion ladder specified in Section VI.A, measuring the score against recursion depth, is the primary direction for future work.

Fifth, the calibration is empirical rather than parametric. Bootstrap thresholding over a disjoint calibration pool holds close to nominal across domains; the naive fit-pool quantile does not, and we neither test multivariate normality on the reference sample nor implement the parametric Hotelling's T-squared branch noted in Section V, so we cannot say whether a distribution-aware threshold would calibrate better or worse. Two defects surfaced during that work, reference-fit leakage and a covariance conditioned at 1.44e12 before shrinkage. Both are fixed, and both are a caution about how much of a small-sample result can be implementation rather than signal.

Sixth, every experiment is small: 36 corpus instances for severity ranking, 12 documents per class across two seeds for the generator folds, 30 test documents per domain across three seeds for false positives, 37 shards for pruning. We report no confidence intervals because these sample sizes do not support them, and any difference of a few hundredths in Section VII should be read as unresolved rather than as a finding.

Seventh, the perplexity feature depends on a specific panel. Independence from the generator is not independence from all models, and we have not run the ablation that would show the score survives a change of panel across architecture, tokenizer, size, and member count. Until that exists, the model-agnostic claim covers the generator only.

Eighth, the paper is English-only, and every domain corpus except source code is template-generated rather than naturally occurring. Real legal, clinical, and encyclopedic text is more heterogeneous than any template, and type-token ratio and rank-frequency slope behave differently across morphologies and scripts. The false-positive study also omits text that is low-diversity by necessity, including writing by non-native speakers and plain-language or accessibility material, where a false positive falls on populations already thin in training corpora. Pruning can narrow topical and demographic coverage, and a deployment should watch what it removes, not only what accuracy it recovers.

The obvious extension is continuous monitoring instead of one-time screening. Corpora get appended to, and a score computed at ingest goes stale; treating contamination as a monitored signal with alerting rather than a gate evaluated once follows what reliability monitoring already does elsewhere [55].

SynthSentry has to live somewhere in a pipeline that already exists. Production pretraining stacks run near-duplicate detection and a perplexity or classifier-based quality filter, and both overlap with what we measure: deduplication strips some of the repetition that drives lexical-diversity collapse, and a quality filter already computes one of the panel's inputs. We place the score after deduplication, so duplicate text does not dominate the tail-truncation feature, and beside quality filtering rather than in its place, because a fluent, high-quality, thoroughly synthetic corpus is exactly what a quality filter waves through and this score is meant to stop. How much contamination signal survives aggressive deduplication is an experiment we run, not an assumption we make.

## IX. Conclusion

The measured results are mixed, and at a scale that settles nothing on its own. Severity ranking works, though on this three-feature problem the composites do not beat lexical diversity alone. Generator-family holdout produces small, mixed-sign deltas that two seeds cannot separate from noise. Per-domain calibration holds its budget once shrinkage and a bootstrap threshold replace the naive quantile. The downstream pruning experiment did not reproduce a contamination-driven accuracy deficit at all, so it cannot speak to recovery, and what it does show is the over-pruning risk we already flagged.